\documentclass[conference]{IEEEtran}
\IEEEoverridecommandlockouts
\usepackage{cite}
\usepackage{amsmath,amssymb,amsfonts}
\usepackage{algorithmic}
\usepackage{graphicx}
\usepackage{textcomp}
\usepackage{xcolor}
\usepackage{makecell}
\usepackage{placeins}   
\usepackage{threeparttable}   

\def\BibTeX{{\rm B\kern-.05em{\sc i\kern-.025em b}\kern-.08em
    T\kern-.1667em\lower.7ex\hbox{E}\kern-.125emX}}
\begin{document}

\title{Mapping Tire-Road Friction Coefficients Along Driven Trajectories \
\thanks{This work was partially funded by Nokia Foundation under grant number 20250107.}
}

\author{\IEEEauthorblockN{
Risto Ojala}
\IEEEauthorblockA{\textit{Energy and Mechanical Engineering} \\
\textit{Aalto University}\\
Espoo, Finland \\
risto.ojala@aalto.fi}

}

\maketitle

\begin{abstract}
Information of the tire-road friction coefficient is critical for intelligent vehicle systems, as friction establishes the limits at which the vehicle can be manoeuvred without loss of control.
While there is extensive literature on online friction estimation, no previous studies have explored post-processing and mapping friction coefficients along a driven trajectory based on recorded sensor data.
Mapped friction information could be shared by connected vehicles, enabling proactive measures in low-friction areas.
This paper presents an offline post-processing method, FrictionMap, for mapping friction coefficients from recorded driving data.
The method operates on commonly available on-board vehicle sensor information, and requires only a basic vehicle parameter set and a slip threshold as prior knowledge.
FrictionMap runs an extended Kalman filter combined with a Rauch-Tung-Striebel smoother to estimate vehicle dynamics.
Data points where the friction limit is observable are identified by filtering points based on tire slip.
A Gaussian process is fitted to these points to acquire a continuous estimate of the friction coefficient along the trajectory.
FrictionMap is validated via a simulation study, and shown to outperform commonly used online friction estimation techniques.
The code implementation and the simulation benchmark are released as open-source after paper acceptance to enable further research.

\end{abstract}

\begin{IEEEkeywords}
Friction estimation, intelligent vehicles, safety, vehicle dynamics
\end{IEEEkeywords}

\section{Introduction}
Estimation of the maximum available tire-road friction with on-board vehicle sensors has been a long-standing estimation problem in vehicle engineering \cite{acosta2017road, liu2025survey, wang2022tire}.
Friction may drastically vary in different weather conditions, especially in colder climates with snow and ice.
For reliable and safe vehicle control, knowing the tire-road friction coefficient is critical for understanding the limits of vehicle manoeuvring.
With the increasing adoption of autonomous driving, advanced driver assistance systems, and intelligent active safety features, there is a concrete need for vehicle systems to more accurately assess the available friction potential.

Commonly, friction estimation has been performed by applying online estimators based on a dynamic vehicle model and measurement signals such as acceleration and wheel rotational speeds.
However, the performance of these methods is inherently limited, as the friction potential can generally be only observed at the limits of the tire grip.
To avoid the problems of online friction estimation, recently there has been increasing attention towards mapping and sharing of friction information \cite{gao2023boxes}.
Connected vehicles have the potential to share their observed friction information with others, maintaining a database of the friction coefficients in differing parts of the road network.
This enables predictive planning and control, as vehicles are aware of the friction conditions ahead of time.
However, these applications rely on accurate friction information being available from the connected vehicles.

\begin{figure}
    \centering
    \includegraphics[width=0.95\columnwidth]{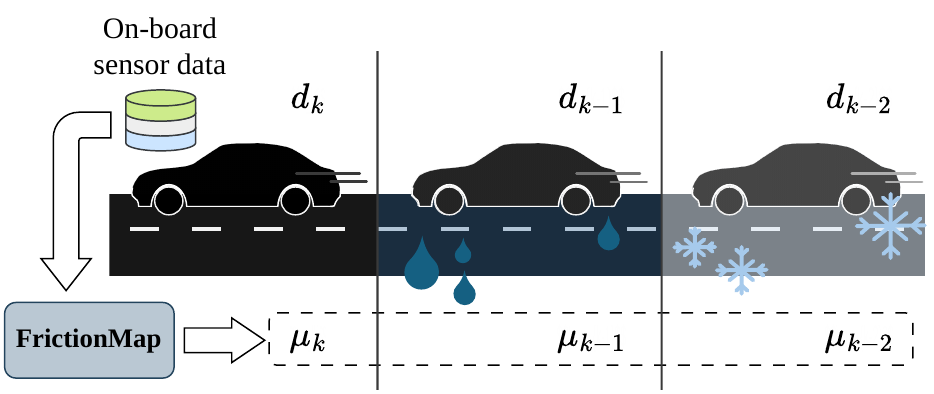}
    \caption{High-level illustration of the proposed FrictionMap solution. Given recorded vehicle on-board sensor data as an input, the solution estimates the tire-road friction coefficient $\mu$ as a function of the distance $d$ along the driven route.}
    \label{fig:graph_abstract}
\end{figure}

This paper studies the mapping of friction coefficient values along a driven vehicle trajectory, proposing the FrictionMap post-processing method for the task.
The general operation principle of FrictionMap is illustrated in Fig. \ref{fig:graph_abstract}.
Based on recorded information of commonly used on-board automotive sensors, the method models the friction coefficient as a function of the driven distance.
The approach is designed to depend on a minimal amount of known vehicle and tire parameters.
We demonstrate that the offline processing of FrictionMap outperforms commonly applied online methods in the friction estimation task.

The main contributions of this work are summarized as follows:

\begin{itemize}
    \item First paper studying post-processing of friction estimates from recorded vehicle on-board sensor data of a driven trajectory.
    \item A novel offline friction estimation method, FrictionMap, which outperforms common online methods.
    \item A novel open-source simulation dataset for reproducibly benchmarking friction estimation solutions.
    
\end{itemize}

\section{Related work}
On-board friction estimation for production vehicles has been studied with a number of different approaches.
Multiple reviews for the field are available, covering the development of the past decades \cite{acosta2017road, liu2025survey, wang2022tire}.
Considering the scope of this study, we focus on approaches based on dynamics modelling and state estimation, and skip the review of remote sensing \cite{ma2022current} and data-driven methods \cite{wang2022tire, schafke2023transformer}.

A common problem with dynamics-based friction estimation is that friction is weakly observable until the peak tire force has been reached.
One proposed strategy to mitigate this issue has been to observe the slip-slope of the tire in low-excitation conditions \cite{gustafsson1997slip}.
The slip-slope has been shown to vary based on the road surface condition, and this information can be used to classify the conditions and approximate a friction coefficient \cite{gustafsson1997slip, rajamani2010tire, rajamani2011algorithms}.
However, as the estimated slip-slope tends to be noisy, reliable estimation without sufficient excitation still remains difficult \cite{acosta2017road, rajamani2010tire}.
Also, the slip-slope may not always correlate with the friction coefficient on different surfaces, leading to incorrect estimates \cite{andrieux2010new}.
Majority of previous approaches have therefore not relied on this estimation strategy.

A common strategy has been to jointly estimate the vehicle dynamics and the friction coefficient in the vehicle state estimation, using a tire model to estimate the friction coefficient.
Wielitzka et al. \cite{wielitzka2017state} studied this approach with an unscented Kalman filter (UKF) and the Pacejka tire model.
In their further studies they explored the credibility of the approach \cite{wielitzka2019credibility} and methods for reducing parameter drift \cite{wielitzka2018sensitivity}.
Wang et al. \cite{wang2020novel} used a cubature Kalman filter with a Pacejka tire model for estimating the friction coefficient based on acceleration measurements.
An adaptive process noise was applied to the friction coefficient estimate, based on the measured acceleration.
Some studies have explored joint estimation with advanced sensor instrumentation for enhanced accuracy.
Lampe et al. \cite{lampe2023analysis} applied a UKF for the estimation task, including measurements from strain gauges installed on the tie rods and potentiometers installed on the springs of the vehicle.

Detaching the friction estimation from the vehicle dynamics estimation has been an alternative design choice.
Ray \cite{ray1997nonlinear} used an extended Kalman filter (EKF) for estimating the vehicle state and tire forces.
The current friction coefficient was identified with Bayesian hypothesis selection, using a Szostak tire model to analyze the forces.
Villagra et al. \cite{villagra2011diagnosis} derived vehicle dynamics analytically from kinematics measurements, and computed tire forces with a weighted Dugoff model.
The friction coefficient was estimated by detecting the saturation of the tire force.
Wan et al. \cite{wan2025enhanced} used analytical derivations and a sliding mode observer for computing the vehicle dynamics.
A square root cubature Kalman filter was deployed to estimate the friction coefficient, using a Brush tire model as the measurement update.
Huang et al. \cite{huang2019calculation} and Li et al. \cite{li2024capturing} applied Kalman filter-based approaches, where the friction coefficient was the only component in the state. 
Updates were propagated through measurements by analytical computation of tire forces.


A popular design choice for detaching the friction estimation from dynamics estimation has been the utilization of a cascaded Kalman filter architecture, consisting of multiple consecutive Kalman filters.
Qi et al. \cite{qi2015estimation} used a cascaded EKF, with the first EKF estimating vehicle dynamics.
The second EKF estimated the friction coefficient, using a modified Dugoff model to perform the measurement update.
Yoon et al. \cite{yoon2016estimation} used a total of three cascaded filters in their approach.
They used the first Kalman filter to estimate yaw and roll angles, the second to estimate slip, and finally a separate EKF to estimate friction based on a brush tire model.
Hu et al. \cite{hu2020real} developed another cascaded approach with also three filters in total. 
An EKF was used to estimate the tire vertical forces and slip, with a parallel UKF estimating the horizontal tire forces.
A second UKF combined the results and estimated the friction coefficient based on the tire forces.

Despite the previous extensive literature on online friction estimation, no previous papers have studied the offline mapping of friction coefficients along a driven trajectory based on vehicle on-board data.
We fill this research gap by proposing the FrictionMap model for the task, demonstrating that the post-processing approach results in more accurate estimation compared to common online methods.
We also share our simulation dataset and code implementation, providing the first open-source benchmark for friction estimation.

\section{Methods}
\subsection{Problem Statement}

The analyzed task is to estimate the friction coefficient $\mu$ as a function of the distance $d$ along a route driven by the vehicle.
As an input, we assume recorded on-board measurement data of commonly available sensors, and a basic known dynamic model of the vehicle.

\subsection{State Equations for Vehicle Dynamics}
\label{sec:dynamics_filtering}
For modelling vehicle dynamics, we use a 2D dynamic bicycle model.
We opt for a simple model in order to promote generalizability and avoid dependency on calibrated parameters for describing e.g. suspension characteristics.
The modelled vehicle is a front-wheel drive, corresponding to the vehicle used in the experiments.
The vehicle model is deployed in an EKF to track the vehicle state.
The vehicle state $\textbf{x}$ is defined as
\begin{equation}
\begin{split}
\mathbf{x} = \bigl[\, & X, Y, \theta, v_x, v_y, r, \omega_f, \omega_r, \\
& F_{x,f}, F_{y,f}, F_{x,r}, F_{y,r}, T_w, \delta \,\bigr],
\end{split}
\end{equation}
where $X$ and $Y$ represent the vehicle coordinates in the global coordinate frame, $\theta$ is the vehicle yaw angle, $v_x$ is vehicle longitudinal speed, $v_y$ is vehicle lateral speed, $r$ is yaw rate, and $T_w$ is the torque delivered by the engine to the front wheels.
The dynamic bicycle model lumps the right and left wheel together in the centre of the axle.
Longitudinal and lateral forces acting on the front wheel are represented by $F_{x,f}$ and $F_{y,f}$, respectively.
Similarly, longitudinal and lateral force on the rear wheel are represented by $F_{x,r}$ and $F_{y,r}$, respectively.
Rotational speeds of the front and rear wheels are denoted as $\omega_f$ and $\omega_r$, respectively. 
The steering angle is denoted by $\delta$.

The vehicle states are modelled with the following equations:
\begin{equation}
\dot X = v_x\cos\theta - v_y\sin\theta, \quad \dot Y = v_x\sin\theta + v_y\cos\theta,
\end{equation}
\begin{equation}
\dot v_x =
\frac{1}{m}
\left(
F_{x, f}\cos\delta - F_{y, f}\sin\delta + F_{x, r}
\right)
+ r v_y,
\end{equation}
\begin{equation}
\dot v_y =
\frac{1}{m}
\left(
F_{y, f}\cos\delta + F_{x, f}\sin\delta + F_{y, r}
\right)
- r v_x,
\end{equation}

\begin{equation}
\dot r =
\frac{1}{I_z}
\left(
l_f (F_{y, f}\cos\delta + F_{x, f}\sin\delta) - l_r F_{y, r}
\right),
\end{equation}
\begin{equation}
\dot\omega_f =
\frac{1}{J_w}
\left(
T_w - R F_{x,f}
\right), \quad \dot\omega_r =
\frac{1}{J_w}
\left(-R F_{x,r}
\right).
\end{equation}




Here, $J_w$ is the wheel inertia, $m$ is the vehicle mass, $R$ is the dynamic radius of the tires, $I_z$ is the vehicle yaw inertia, $l_f$ is the distance between the center of gravity (CoG) and the front axle, and $l_r$ is the distance between the CoG and the rear axle.
The horizontal tire forces, the torque on the wheels, and the steering angle are modelled with Gaussian random walk.
The derivative of yaw angle is modelled as the yaw rate $\dot\theta=r$.

For the EKF update step, the following measurement vector $\textbf{z}$ and measurement model $h(\textbf{x})$ are used:
\begin{equation}
\begin{split}
\mathbf{z} = \bigl[\, & X_{\mathrm{GNSS}}, Y_{\mathrm{GNSS}}, \dot X_{\mathrm{GNSS}}, \dot Y_{\mathrm{GNSS}}, a_{x,\mathrm{IMU}}, a_{y,\mathrm{IMU}}, \\
& r_{\mathrm{IMU}}, \omega_{f,\mathrm{enc}}, \omega_{r,\mathrm{enc}}, T_{e,\mathrm{ECU}}, \delta_{\mathrm{SAS}} \,\bigr],
\end{split}
\end{equation}
\begin{equation}
\begin{split}
h(\mathbf{x}) = \bigl[\, & X, Y, \dot X, \dot Y, \dot v_x - r v_y, \dot v_y + r v_x, \\
& r, \omega_f, \omega_r, T_w / i_s, \delta \,\bigr].
\end{split}
\end{equation}
The measurements are available on most modern vehicles through the Global Navigation Satellite System (GNSS), inertial measurement unit (IMU), the wheel encoders (enc), the electronic control unit (ECU) and the steering angle sensor (SAS).
Longitudinal and lateral accelerations are denoted by $a_x$ and $a_y$, respectively.
The engine torque is denoted by $T_e$ and the gear ratio between the engine and the wheels is denoted by $i_s$.
The combined wheel rotation speeds for the bicycle model are acquired by averaging the left and right wheel speeds.

Based on the estimated states, tire slip is evaluated analytically. 
Longitudinal slip $\kappa$ and lateral slip $\alpha$ are computed for a wheel as:
\begin{equation}
\kappa =
\frac{R\omega - v_{x,w}}{|v_{x,w}|}, \quad
\alpha =
-\tan^{-1}
\left(
\frac{v_{y,w}}{v_{x,w}}
\right).
\end{equation}

Here, $v_{x,w}$ is longitudinal speed of the wheel, $v_{y,w}$ is the lateral speed of the wheel.
For the combined front wheel of the bicycle model, these speeds are computed as:
\begin{equation}
v_{x,w,f} = v_x \cos \delta + (v_y + l_f r) \sin \delta,
\end{equation}
\begin{equation}
v_{y,w,f} = -v_x \sin \delta + (v_y + l_f r) \cos \delta.
\end{equation}
For the combined rear wheel, these are
\begin{equation}
v_{x,w,r} = v_x, \quad v_{y,w,r} = v_y - l_r r.
\end{equation}

Additionally, longitudinal weight shift is modelled based on the states to more accurately assess the friction coefficient.
The normal load on the front wheel $F_{z,f}$ and the rear wheel $F_{z,r}$ are defined as:
\begin{equation}
F_{z,f} = F_{z,f,s}-\frac{m(\dot v_x-rv_y)h_{\mathrm{CoG}}}{l_f+l_r},
\end{equation}
\begin{equation}
F_{z,r} = F_{z,r,s}+\frac{m(\dot v_x-rv_y)h_{\mathrm{CoG}}}{l_f+l_r},
\end{equation}
where $F_{z,f,s}$ and $F_{z,r,s}$ are the static normal load on the combined front and rear axles, respectively.
The described vehicle modelling provides the foundation on which the friction estimation methodology of this paper is built.

\subsection{The Proposed Algorithm: FrictionMap}
\begin{figure*}
    \centering
    \includegraphics[width=0.95\textwidth]{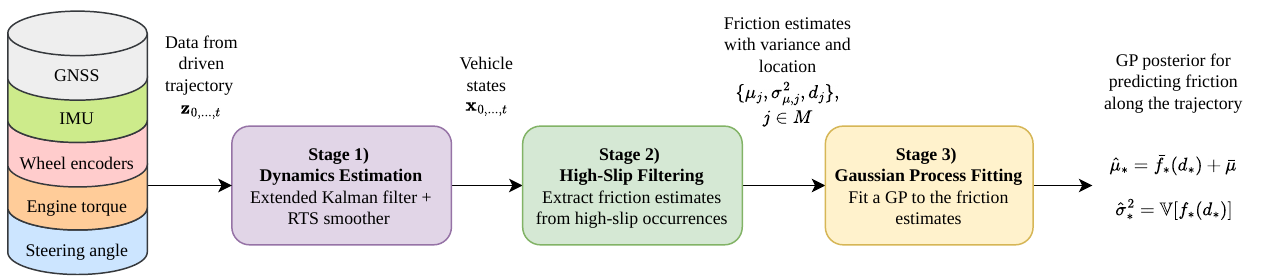}
    \caption{Overview of the proposed friction mapping pipeline, FrictionMap.}
    \label{fig:algorithm}
\end{figure*}

We propose the FrictionMap method for estimating the friction coefficient along a driven route, based on recorded vehicle data.
The algorithm overview is given in Fig. \ref{fig:algorithm}, with the details for each of the algorithm stages described below. 

\textbf{Stage 1) Dynamics Estimation.} 
The EKF described in Section \ref{sec:dynamics_filtering} is used to process the sensor information.
Additionally, a Rauch-Tung-Striebel (RTS) smoother is run on the results of the EKF, to further enhance the accuracy of the state estimation.
This results in acquisition of the vehicle state $\textbf{x}_t$ for each timestep $t$.
Based on the acquired states, the longitudinal $\kappa_t$ and lateral tire slips $\alpha_t$ are computed for both the front and rear wheels.
The estimated states and tire slips are passed on into the next stage.

\textbf{Stage 2) High-Slip Filtering.} 
The friction coefficient can be identified by analyzing the peak tire forces during high-slip events.
Therefore, we run a high-slip filter to focus our estimation on data points where the tire forces reach the limits of the grip.
The filtering is carried out on both the front and the rear tires.
The values from Stage 1 are thresholded with a slip limit $\tau$, chosen based on prior knowledge of the tire properties.
This generates a set of candidate indices $H$ as
\begin{equation}
H = \left\{ t \;\middle|\; \sqrt{\kappa_t^2 + (\tan\alpha_t)^2} > \tau \right\}.
\end{equation}
Since the tire forces can be expected to reach the grip limits at these datapoints, these indices are used to generate a set of candidate friction coefficients as
\begin{equation}
\mu_j = \frac{\sqrt{F_{x,j}^2 + F_{y,j}^2}}{F_{z,j}}, \quad j \in H.
\end{equation}
The EKF+RTS smoothed tire force covariances are also used to linearly propagate a variance $\sigma_\mu^2$ for each candidate friction coefficient as
\begin{equation}
\sigma_\mu^2 = \mathbf{J}_\mu \boldsymbol{\Sigma}_{F_x,F_y} \mathbf{J}_\mu^T
\end{equation}
where $\mathbf{J}_\mu = \begin{bmatrix}\frac{\delta\mu}{dF_x} \frac{\delta\mu}{dF_y} \end{bmatrix}$ and $\boldsymbol{\Sigma}_{F_x,F_y}$ is the EKF covariance estimate for $F_x$ and $F_y$. 
Distances along the route $d_j$ corresponding to the friction coefficient candidates are known via the $X_j$ and $Y_j$ state estimates.
A tire experiencing high slip produces a number of data points in a specific location.
To identify the friction coefficient, corresponding to the peak tire grip, a spatial filter is run to acquire descriptive predictions.
The friction coefficient candidates from both the front and the rear tire are appended to a common set, and the driven route is grouped into spatial bins of equal length, $p$.
From each bin, the $n$ highest friction coefficient candidates are taken and appended into a final prediction set $M$.

The final predictions are sparse data points featuring the friction coefficient, the variance, and the corresponding location
\begin{equation}
\{\mu_j, \sigma^2_{\mu,j}, d_j\}, \quad j \in M.
\end{equation}
These sparse points are used to generate a continuous friction estimate in the next processing stage.

\textbf{Stage 3) Gaussian Process Fitting.}
A Gaussian process (GP) is used for modelling the friction coefficient based on the data points from the previous stage.
The friction values $\mu_j$ are centered based on the sample mean $\bar\mu$ as $\mu_j' = \mu_j - \bar\mu$ and the friction estimation function $f(d)$ is given a prior
\begin{equation}
f(d) \sim \mathcal{GP}(0, k(d, d')),
\end{equation}
where $\mathcal{GP}$ denotes the Gaussian process with zero mean and $k(d, d')$ as the covariance function.
We choose a Mat\'ern kernel as the covariance function
\begin{equation}
k(d, d') = \sigma_k^2\frac{2^{1-\nu}}{\Gamma(\nu)}
\left( \frac{\sqrt{2\nu}\, |d - d'|}{\ell} \right)^{\nu}
K_{\nu}\!\left( \frac{\sqrt{2\nu}\, |d - d'|}{\ell} \right),
\end{equation}
where $\sigma_k^2$ is the kernel variance, $\ell$ denotes length scale, $\nu$ is the smoothness parameter, $\Gamma(\cdot)$ is the gamma function and $K_{\nu}(\cdot)$ is the modified Bessel function of the second kind.
Each friction observation from the final prediction set $M$ of Stage 2 is modelled as
\begin{equation}
\mu_j' = f(d_j) + \epsilon_j, \quad \epsilon_j \sim \mathcal{N}(0,\sigma^2_{\mu,j}).
\end{equation}
The kernel parameters are optimized by maximizing the log-marginal likelihood, and the GP posterior is computed \cite{pedregosa2011scikit, rasmussen2003gaussian}.
The GP posterior is used to generate the final friction estimates as mean $\hat\mu_*$ and variance $\hat\sigma^2_*$ for query distances $d_*$ as
\begin{equation}
\hat\mu_* = \bar{f_*}(d_*) + \bar\mu,
\end{equation}
\begin{equation}
\hat\sigma^2_* = \mathbb{V}[f_*(d_*)],
\end{equation}
where $\bar{f_*}(d_*)$ and $\mathbb{V}[f_*(d_*)]$ are the mean and variance of the GP posterior \cite{rasmussen2003gaussian}, respectively.
The sample mean $\bar\mu$ is used to shift the estimate to represent real friction values.
This GP posterior is used for producing the final friction estimates, with uncertainty, along the driven trajectory.

\begin{figure*}
    \centering
    \includegraphics[width=0.95\textwidth]{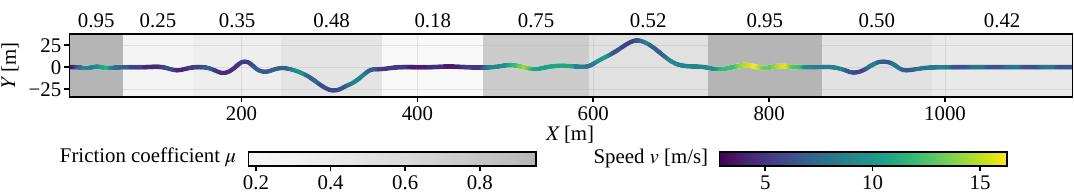}
    \caption{Driven trajectory of the simulation dataset, with vehicle speed and friction patches visualized. Values above the plot indicate the exact friction coefficient at each patch.}
    \label{fig:trajectory_speed}
\end{figure*}

\subsection{Baselines}
To compare FrictionMap with reasonable baselines, two different online EKF-based estimators were used for benchmarking.
Both estimators were implemented with two different tire models: a linear model and a combined-slip Dugoff model with weighting similar to \cite{villagra2011diagnosis}.
The initial slope of the tire models was matched to the ground truth tire model, and the weighted Dugoff model was parametrized to intersect the ground truth tire model at the peak of the lateral force-slip curve.
To factor in the limitations of the available friction, combining both lateral and longitudinal limits, the tire model force estimates were saturated with a Coulomb friction circle assumption.
Simple tire models were chosen for the baselines, as more advanced models are practically difficult to apply in production vehicles, as they require detailed calibration.
This is in line with the prior requirements of FrictionMap, which assumes only rough knowledge of the tire grip limits.

\textbf{Joint EKF}. The first baseline method includes the friction coefficient in the state estimation, using a joint EKF for estimating both dynamics and the friction.
Joint estimation has been previously applied in \cite{wielitzka2017state, wang2020novel}.
The same EKF described in Section \ref{sec:dynamics_filtering} is used, except for the following differences.
The state is appended with the friction coefficient, with the prediction step of the coefficient modelled as Gaussian random walk.
Additionally, the prediction step of the tire forces is modified to use the forces estimated by a tire model.
The prediction step for each of the tire forces $F \in \{F_{x,f}, F_{y,f}, F_{x,r}, F_{y,r}\}$ is modelled as:
\begin{equation}
\dot F = \frac{F_{TM} - F}{\Delta t},
\end{equation}
estimating the derivative by using both the tire model estimate and the current force state estimate.
The timestep between samples is denoted by $\Delta t$, and the force estimated by the tire model is denoted by $F_{TM}$.

\textbf{Cascaded EKF.} The second online baseline estimator uses a cascaded EKF.
Cascaded estimation has been previously applied in \cite{qi2015estimation, yoon2016estimation}.
The first EKF is identical to the definition in Section \ref{sec:dynamics_filtering}, estimating the vehicle dynamics.
The subsequent EKF estimates the friction coefficient $\mu$ based on this information.
The state in the second EKF is
\begin{equation}
\textbf{x}_c = \begin{bmatrix} \mu \end{bmatrix},
\end{equation}
with the prediction step being modelled as Gaussian random walk.
The measurement vector and measurement model are given as
\begin{equation}
\mathbf{z}_c =
\begin{bmatrix}
F_{x,f}, F_{y,f}, F_{x,r}, F_{y,r}
\end{bmatrix},
\end{equation}
\begin{equation}
h_c(\textbf{x}_c)=
\begin{bmatrix}
F_{x,f, TM}, F_{y,w_f, TM}, F_{x,w_r,TM}, F_{y,w_r,TM}
\end{bmatrix},
\end{equation}
using the force estimates from the tire model to update the friction coefficient estimate.

\textbf{Adaptive Variants.} Previous research has explored adaptive Kalman filters to combat the problem of the friction estimate drifting due to low excitation \cite{li2024capturing,wang2020novel}.
Adaptive versions of the baselines were also implemented, using the slip uncertainty during each timestep to estimate when the friction coefficient is realistically observable.
Slip $\textbf{s} = \begin{bmatrix} \kappa_f, \kappa_r, \alpha_f, \alpha_r \end{bmatrix}$ is computed based on the state variables $\textbf{v}_s = \begin{bmatrix} v_x, v_y, r, \omega_f, \omega_r, \delta\end{bmatrix}$, and therefore the slip covariance can be linearly propagated as 
\begin{equation}
\boldsymbol{\Sigma}_s= \textbf{J}_s\boldsymbol{\Sigma}_{\textbf{v}_s}\textbf{J}_s^T,
\end{equation}
where $\textbf{J}_s = \frac{\delta\textbf{s}}{\delta\textbf{v}_s}$ and $\boldsymbol{\Sigma}_{\textbf{v}_s}$ is the covariance of state variables $\textbf{v}_s$ from the EKF estimate.
When running the adaptive filters, if both $|\alpha| < 3\sigma_\alpha$ and $|\kappa| < 3\sigma_\kappa$ for a tire, the friction observations of the tire are considered uninformative.
For the Joint EKF, this results in the force predictions being swapped to random walk instead of using the tire model.
For the Cascaded EKF, this results in the measurement covariance of the tire model-based update step being increased to $10^{18}$.
These steps effectively decouple the tire model from the friction estimation, preventing drift caused by noisy estimates.
When either the lateral or longitudinal absolute slip values exceed their respective threshold, the filters operate normally without any modification.

\subsection{Simulation Dataset}
Project Chrono \cite{tasora2015chrono, serban2019chrono} version 9.0.1 was used for creating a simulation dataset for the friction mapping task.
The default Sedan model from the Chrono::Vehicle library of the project was used in the simulation, with the tires being modelled with a TMEasy tire model.
The lateral maximum grip limit of the TMEasy model at the static vehicle normal load was used as the reference friction coefficient.
This approximation was necessary, as the estimation task does not account for the differences in the lateral and longitudinal grip.
The parameter values used for modelling the Sedan are provided in Table \ref{tab:variables}.

\begin{table}
\caption{Values used for the model parameters.}
\label{tab:variables}
\centering
\begin{tabular}{c c c}
\hline
\textbf{Variable} & \textbf{Definition} & \textbf{Value} \\
\hline
$m$ & vehicle mass & 1669 kg \\
$I_z$ & vehicle yaw inertia & 1356 $\text{kgm}^2$ \\
$J_{w}$ & combined wheel inertia & 2.198 $\text{kgm}^2$ \\
$l_{f}$ & distance from CoG to front axle & 1.388 m \\
$l_{r}$ & distance from CoG to rear axle & 1.388 m \\
$R$ & tire dynamic radius & 0.3236 m\\
$F_{z,f,s}$ & static front axle load & 8186 N \\
$F_{z,r,s}$ & static rear axle load & 8186 N \\
$h_{\mathrm{CoG}}$ & height of the centre of gravity & 0.46 m \\
$i$ & engine-to-wheel gear ratios & \makecell[t]{\{18.89, 10.23, 6.38, \\ 4.70, 3.92, 3.34\}} \\ 
\hline
\end{tabular}
\end{table}

The simulation consisted of the Sedan vehicle driving on a flat, level surface with varying friction coefficients.
Waypoints were established through the friction patches, and the vehicle was controlled by the Project Chrono path follower.
The controller was supplemented with basic anti-lock braking and traction control, limiting the braking or acceleration if longitudinal slip reached absolute values higher than 0.4.
The route was designed to feature sharp accelerations and decelerations, as well as pronounced steering, to provide sufficient lateral and longitudinal excitation.
This was a conscious design choice, as the analyzed friction estimation approaches require excitation near the limits of the grip for predicting the friction coefficient.
The driven route is visualized in Fig. \ref{fig:trajectory_speed}, showing both the speed of the vehicle and the friction patches along the route.

\begin{figure}
    \centering
    \includegraphics[width=0.95\columnwidth]{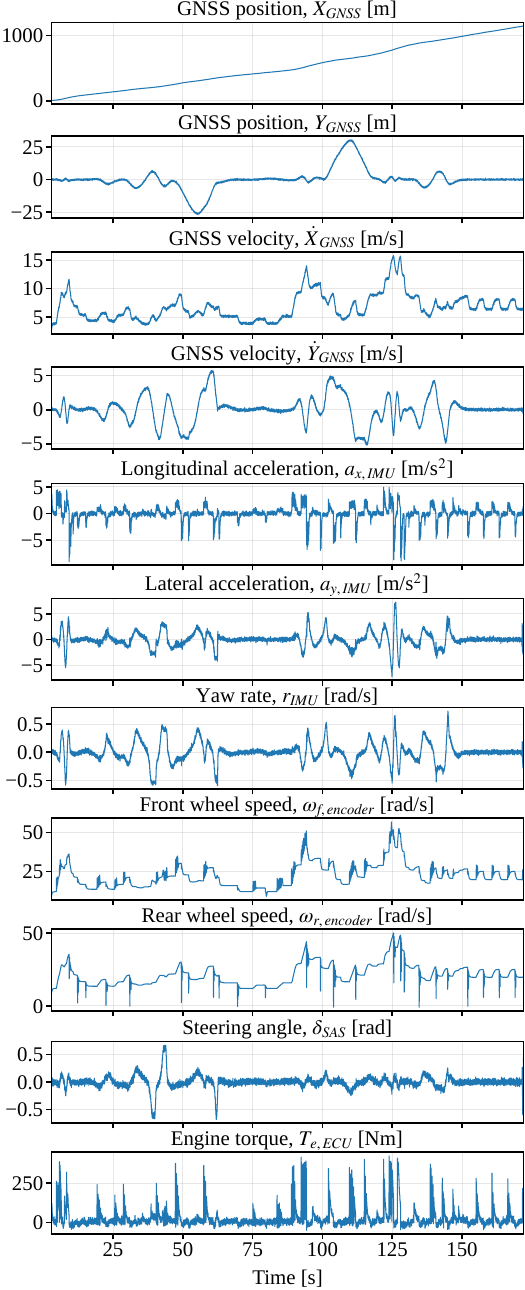}
    \caption{Noisy measurement signals of the simulation dataset.}
    \label{fig:noisy_signals}
\end{figure}

To acquire the required sensor signals for the estimation, ground truth values were taken from the simulation environment and white noise was added to the readings to simulate sensor data.
The following standard deviations were used when adding the noise: 0.3 m for GNSS positioning, 0.1 m/s for GNSS velocity, 0.02 rad/s for the IMU yaw rate, 0.2 m/s$^2$ for the IMU accelerations, 0.05 rad/s for the wheel encoders, 0.03 rad for the steering angle, and 10 Nm for the engine torque.
The sensor readings are visualized in Fig. \ref{fig:noisy_signals}.
The kinematic sensor readings were logged with respect to the CoG of the vehicle.

\section{Experiments}

\begin{figure*}
    \centering
    \includegraphics[width=0.95\textwidth]{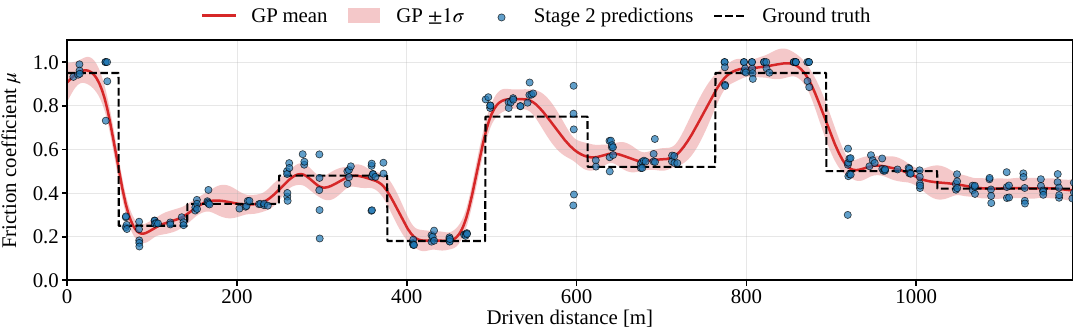}
    \caption{Friction estimates of the FrictionMap Gaussian process posterior compared against the ground truth reference values.}
    \label{fig:frictionmap_result}
\end{figure*}

\subsection{Implementation Details}
FrictionMap and the baselines were implemented in Python, using the FilterPy-library for the extended Kalman filter, and scikit-learn for the GP implementation.
In FrictionMap, the slip threshold $\tau$ was set to 0.15, the binning distance $p$ was set to 20 m and the number of values in a bin $n$ was set to 5.
The Matérn-kernel smoothness $\nu$ was set to 1.5, the kernel variance $\sigma_k^2$ was optimized from a range of $[0.01,10]$ and the length scale $l$ was optimized from a range of $[10,500]$.
The EKF in FrictionMap was used with a diagonal process noise and measurement noise matrices.
The process noise standard deviations were 0.1\,m for position, 0.1\,m/s for velocity, 0.1\,rad/s for yaw rate, 0.01\,rad for both heading and steering angle, 10\,rad/s for wheel speeds, 10\,Nm for wheel torque and 1\,kN for tire forces.
The measurement noise standard deviations were 0.1\,m for GNSS position, 0.5\,m/s for GNSS velocity, 0.1\,m/s$^2$ for IMU acceleration, 0.1\,rad/s for IMU yaw rate, 0.1\,rad/s for wheel encoders, 10\,Nm for engine torque and 0.02\,rad for the steering angle. 
Identical values were used in the baselines, with the friction coefficient process noise having a standard deviation of 0.05 and the Cascaded EKF tire model estimates having equal measurement noise as the process noise on the tire forces.
Additionally, all friction estimates from the models were bounded to a range of $[0.1,1]$ to ensure predictions remained within a reasonable range.

\subsection{Friction Estimation Accuracy}
\begin{figure}
    \centering
    \includegraphics[width=0.95\columnwidth]{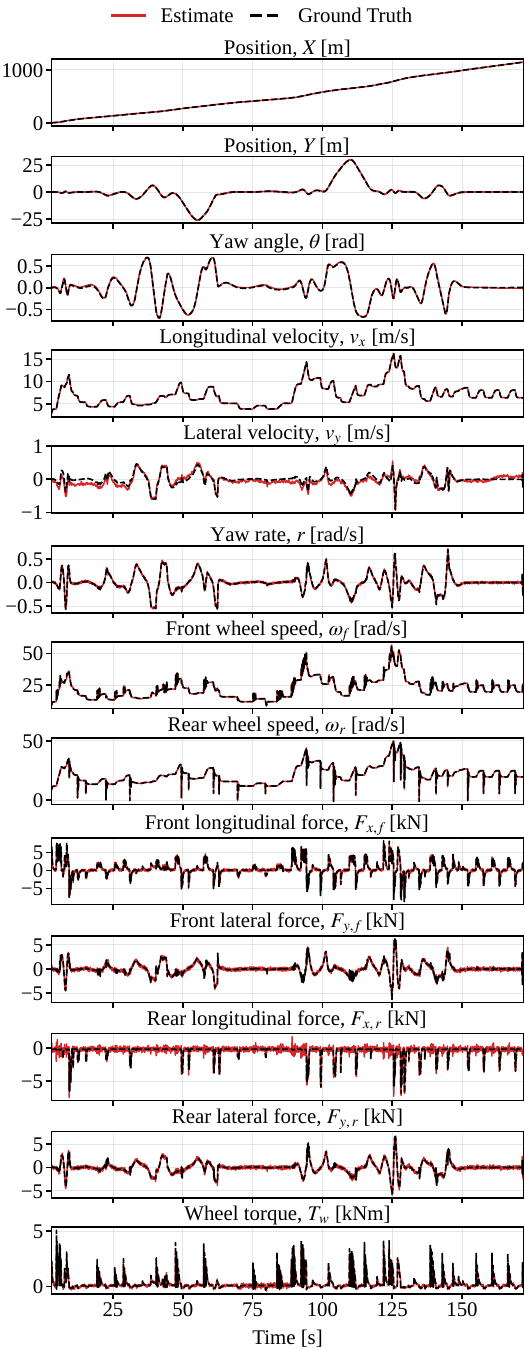}
    \caption{State estimation accuracy of the EKF + RTS smoother in FrictionMap.}
    \label{fig:state_estimation_accuracy}
\end{figure}

The simulation dataset was used for benchmarking the friction estimation accuracy of FrictionMap against the baselines.
The used metrics were mean absolute error (MAE), root-mean squared error (RMSE), and prediction interval coverage probability (PICP).
The results of the benchmark are presented in Table \ref{tab:accuracy}, showing that FrictionMap reaches a notably higher accuracy than the baseline solutions.
Additionally, the confidence intervals of FrictionMap offer a reasonable measure of uncertainty.
These findings indicate that the post-processing approach of FrictionMap provides concrete benefits for the estimation task.
A detailed visualization of FrictionMap estimates on the simulation dataset is shown in Fig. \ref{fig:frictionmap_result}.
The fitted GP posterior follows the ground truth closely, although not perfectly tracking the transition regions due to limited data points.
The state estimation accuracy of the EKF + RTS smoother used in FrictionMap is shown in Fig. \ref{fig:state_estimation_accuracy}.
The lateral velocity estimates can be seen to slightly deviate from the true values, as a consequence of the noisy measurement data.
This results in noise in the slip angle estimates, likely being one of the most notable sources of error in the estimation.

\begin{table}
\caption{Results of the friction estimation accuracy benchmark. Best value in each column is bolded.}
\label{tab:accuracy}
\centering
\begin{threeparttable}
\begin{tabular}{c | c c c}
\hline
\textbf{Method} & \textbf{MAE} & \textbf{RMSE} & \textbf{PICP ($1\sigma$)}\tnote{a} \\
\hline
Joint EKF (linear) & 0.279 & 0.341 & 0.289 \\
Joint EKF (Dugoff) & 0.259 & 0.325 & 0.384 \\
Cascaded EKF (linear) & 0.287 & 0.343 & 0.078 \\
Cascaded EKF (Dugoff) & 0.288 & 0.350 & 0.089 \\
\hline
Joint EKF, adaptive (linear) & 0.143 & 0.186 & 0.883  \\
Joint EKF, adaptive (Dugoff) & 0.136 & 0.179 & 0.900  \\
Cascaded EKF, adaptive (linear) & 0.105 & 0.133 & 0.887 \\
Cascaded EKF, adaptive (Dugoff) & 0.103 & 0.131 & 0.895 \\
\hline
FrictionMap [ours] & \textbf{0.050} & \textbf{0.087} & \textbf{0.754} \\
\hline
\end{tabular}
\begin{tablenotes}
\footnotesize
\item[a] Target = 0.683
\end{tablenotes}
\end{threeparttable}
\end{table}

\subsection{Ablation Study}
An ablation study was performed to analyze the impact of different design choices in the FrictionMap architecture, as well as estimate the sensitivity of the method to different hyperparameter choices.
Results of the ablation study are shown in Table \ref{tab:ablations}.
The ablations included a test with the RTS smoother disabled, and a test where instead of using a GP the friction estimates were produced by nearest neighbor matching to the Stage 2 candidates.
The results show that the changes in the architecture result in reduced accuracy.
The ablated hyperparameters were the $\nu$ value of the GP kernel and the $\tau$ value of the Stage 2 slip threshold.
These ablations resulted in minimal change in the performance, highlighting the robustness of the FrictionMap algorithm.

\begin{table}
\caption{Ablation study of different design choices in the architecture. Best value in each column is bolded.}
\label{tab:ablations}
\centering
\begin{threeparttable}
\begin{tabular}{c | c c c}
\hline
\textbf{Method} & \textbf{MAE} & \textbf{RMSE} & \textbf{PICP ($1\sigma$)}\tnote{a}\\
\hline
FrictionMap w/o RTS smoother & 0.059 & 0.096 & \textbf{0.731} \\
FrictionMap w/o GP & 0.072 & 0.136 & - \\
FrictionMap w/ $\nu = 0.5$ & 0.051 & 0.088 & 0.809 \\
FrictionMap w/ $\tau = 0.10$ & 0.055 & 0.089 & 0.732 \\
\hline
FrictionMap [ours] & \textbf{0.050} & \textbf{0.087} & 0.754 \\
\hline
\end{tabular}
\begin{tablenotes}
\footnotesize
\item[a] Target = 0.683
\end{tablenotes}
\end{threeparttable}
\end{table}

\section{Discussion}
The acquired results indicate that FrictionMap can be used for accurately estimating the friction coefficient along driven trajectories based on commonly available on-board sensor data.
The algorithm operates on minimal required prior knowledge of the vehicle, only using a basic dynamic bicycle model and a slip-threshold for extracting observations of the friction coefficient.
However, the experiments have a number of limitations which cause uncertainty in the results.
The experiments were carried out with simulated data, which results in a domain gap to real-world applications.
The simulations were carried out on a perfectly level and even ground plane, ignoring effects caused by road slope and variation in road surface quality.
Additionally, a relatively simple tire model (TMEasy) was used as the ground truth, ignoring detailed effects of different road conditions on the tire dynamics.

The applicability of FrictionMap for natural real-world driving data also remains uncertain, as the simulated drive cycle featured aggressive driving and artificially abrupt friction coefficient changes.
This was a deliberate design choice, as all the benchmarked methods rely on high-slip excitation for identifying the friction coefficient. 
Therefore, the experiment was designed to ensure that the friction coefficient is observable, to ensure a reasonable comparison of the different methods.
Performance for natural driving data can be expected to degrade, as less aggressive manoeuvring results in fewer observations of the peak tire forces.

Future work should explore the performance of FrictionMap with real-world experiments. 
This would enable assessment of the applicability on natural driving data, as well as enable further development for more accurately mapping road friction.
The simulation dataset of this study is released open-source, and can be used to develop and benchmark alternative approaches for friction estimation.


\section*{Acknowledgment}
Large language models (Claude 5 Sonnet and Opus) were used in the code implementation of the presented methodology, experimental evaluation, and related plots (Sections III-IV).
The authors reviewed all AI-assisted content and take full responsibility for it.



\bibliographystyle{ieeetr}
\bibliography{refs}

@article{ma2022current,
  title={Current non-contact road surface condition detection schemes and technical challenges},
  author={Ma, Yao and Wang, Meizhu and Feng, Qi and He, Zhiping and Tian, Mi},
  journal={Sensors},
  volume={22},
  number={24},
  pages={9583},
  year={2022},
  publisher={Multidisciplinary Digital Publishing Institute}
}

@article{acosta2017road,
  title={Road friction virtual sensing: A review of estimation techniques with emphasis on low excitation approaches},
  author={Acosta, Manuel and Kanarachos, Stratis and Blundell, Mike},
  journal={Applied Sciences},
  volume={7},
  number={12},
  pages={1230},
  year={2017},
  publisher={MDPI}
}

@article{gustafsson1997slip,
  title={Slip-based tire-road friction estimation},
  author={Gustafsson, Fredrik},
  journal={Automatica},
  volume={33},
  number={6},
  pages={1087--1099},
  year={1997},
  publisher={Elsevier}
}

@article{ray1997nonlinear,
  title={Nonlinear tire force estimation and road friction identification: Simulation and experiments},
  author={Ray, Laura R},
  journal={Automatica},
  volume={33},
  number={10},
  pages={1819--1833},
  year={1997},
  publisher={Elsevier}
}

@article{andrieux2010new,
  title={New results on the relation between tyre--road longitudinal stiffness and maximum available grip for motor car},
  author={Andrieux, Arnaud and Vandanjon, Pierre Olivier and Lengell{\'e}, R{\'e}gis and Chabanon, Christian},
  journal={Vehicle System Dynamics},
  volume={48},
  number={12},
  pages={1511--1533},
  year={2010},
  publisher={Taylor \& Francis}
}

@article{rajamani2011algorithms,
  title={Algorithms for real-time estimation of individual wheel tire-road friction coefficients},
  author={Rajamani, Rajesh and Phanomchoeng, Gridsada and Piyabongkarn, Damrongrit and Lew, Jae Y},
  journal={IEEE/ASME transactions on mechatronics},
  volume={17},
  number={6},
  pages={1183--1195},
  year={2011},
  publisher={IEEE}
}

@article{rajamani2010tire,
  title={Tire-road friction-coefficient estimation},
  author={Rajamani, Rajesh and Piyabongkarn, Neng and Lew, Jae and Yi, Kyongsu and Phanomchoeng, Gridsada},
  journal={IEEE Control Systems Magazine},
  volume={30},
  number={4},
  pages={54--69},
  year={2010},
  publisher={IEEE}
}

@article{wang2022tire,
  title={Tire road friction coefficient estimation: Review and research perspectives},
  author={Wang, Yan and Hu, Jingyu and Wang, Fa’an and Dong, Haoxuan and Yan, Yongjun and Ren, Yanjun and Zhou, Chaobin and Yin, Guodong},
  journal={Chinese Journal of Mechanical Engineering},
  volume={35},
  number={1},
  pages={6},
  year={2022},
  publisher={Springer}
}

@article{huang2019calculation,
  title={Calculation algorithm of tire-road friction coefficient based on limited-memory adaptive extended Kalman filter},
  author={Huang, Bin and Fu, Xiang and Wu, Sen and Huang, Song},
  journal={Mathematical Problems in Engineering},
  volume={2019},
  number={1},
  pages={1056269},
  year={2019},
  publisher={Wiley Online Library}
}

@inproceedings{schafke2023transformer,
  title={Transformer neural networks for maximum friction coefficient estimation of tire-road contact using onboard vehicle sensors},
  author={Sch{\"a}fke, Hendrik and Lampe, Nicolas and Kortmann, Karl-Philipp},
  booktitle={2023 62nd IEEE Conference on Decision and Control (CDC)},
  pages={5331--5338},
  year={2023},
  organization={IEEE}
}

@article{liu2025survey,
  title={A survey and comprehensive taxonomy of tire-road adhesion coefficient estimation for intelligent vehicles},
  author={Liu, Jiahui and Liu, Yang and Wang, Liang and Qu, Xiaobo},
  journal={IEEE Transactions on Intelligent Transportation Systems},
  volume={26},
  number={9},
  pages={12819--12833},
  year={2025},
  publisher={IEEE}
}

@article{qi2015estimation,
  title={Estimation of the tyre--road maximum friction coefficient and slip slope based on a novel tyre model},
  author={Qi, Zhiquan and Taheri, Saied and Wang, Baofeng and Yu, Hongxiao},
  journal={Vehicle System Dynamics},
  volume={53},
  number={4},
  pages={506--525},
  year={2015},
  publisher={Taylor \& Francis}
}

@article{villagra2011diagnosis,
  title={A diagnosis-based approach for tire--road forces and maximum friction estimation},
  author={Villagra, Jorge and d’Andr{\'e}a-Novel, Brigitte and Fliess, Michel and Mounier, Hugues},
  journal={Control engineering practice},
  volume={19},
  number={2},
  pages={174--184},
  year={2011},
  publisher={Elsevier}
}

@inproceedings{lampe2023analysis,
  title={Analysis of the potential of onboard vehicle sensors for model-based maximum friction coefficient estimation},
  author={Lampe, Nicolas and Ziaukas, Zygimantas and Westerkamp, Clemens and Jacob, Hans-Georg},
  booktitle={2023 American Control Conference (ACC)},
  pages={1622--1628},
  year={2023},
  organization={IEEE}
}

@article{wan2025enhanced,
  title={Enhanced tire road friction coefficient estimation through interacting multiple model design based on tire force observation},
  author={Wan, Wenkang and Sheng, Kai and Cai, Qing and Ao, Lei and Ouyang, Nan and Feng, Haoxuan and Sime, Dejene M},
  journal={Nonlinear Dynamics},
  volume={113},
  number={13},
  pages={15925--15941},
  year={2025},
  publisher={Springer}
}

@article{li2024capturing,
  title={Capturing rapid vehicle dynamics: A refined adaptive Kalman filter for enhanced tire--road friction estimation},
  author={Li, Bin and Zhang, Lin and Zhao, Chunlai and Lu, Jiaxing and Meng, Qiang and Zhang, Zeyang and Chen, Hong},
  journal={IEEE Transactions on Industrial Electronics},
  volume={71},
  number={11},
  pages={14813--14822},
  year={2024},
  publisher={IEEE}
}

@inproceedings{wang2020novel,
  title={A novel approach for tire-road friction coefficient estimation using adaptive cubature kalman filter},
  author={Wang, Yan and Yin, Guodong and Dong, Haoxuan},
  booktitle={2020 4th CAA International Conference on Vehicular Control and Intelligence (CVCI)},
  pages={30--34},
  year={2020},
  organization={IEEE}
}

@article{hu2020real,
  title={Real-time estimation of tire--road friction coefficient based on lateral vehicle dynamics},
  author={Hu, Juqi and Rakheja, Subhash and Zhang, Youmin},
  journal={Proceedings of the Institution of Mechanical Engineers, Part D: Journal of Automobile Engineering},
  volume={234},
  number={10-11},
  pages={2444--2457},
  year={2020},
  publisher={SAGE Publications Sage UK: London, England}
}

@article{gao2023boxes,
  title={Boxes-based representation and data sharing of road surface friction for CAVs},
  author={Gao, Liming and Mitrovich, Juliette and Beal, Craig and Bai, Wushuang and Maddipatla, Satya Prasad and Chen, Cindy and Jerath, Kshitij and Haeri, Hossein and Sinanaj, Lorina and Brennan, Sean},
  journal={Data Science for Transportation},
  volume={5},
  number={2},
  pages={9},
  year={2023},
  publisher={Springer}
}

@inproceedings{tasora2015chrono,
  title={Chrono: An open source multi-physics dynamics engine},
  author={Tasora, Alessandro and Serban, Radu and Mazhar, Hammad and Pazouki, Arman and Melanz, Daniel and Fleischmann, Jonathan and Taylor, Michael and Sugiyama, Hiroyuki and Negrut, Dan},
  booktitle={international conference on high performance computing in science and engineering},
  pages={19--49},
  year={2015},
  organization={Springer}
}

@article{serban2019chrono,
  title={Chrono:: Vehicle: template-based ground vehicle modelling and simulation},
  author={Serban, Radu and Taylor, Michael and Negrut, Dan and Tasora, Alessandro},
  journal={International Journal of Vehicle Performance},
  volume={5},
  number={1},
  pages={18--39},
  year={2019},
  publisher={Inderscience Publishers (IEL)}
}

@article{pedregosa2011scikit,
  title={Scikit-learn: Machine learning in Python},
  author={Pedregosa, Fabian and Varoquaux, Ga{\"e}l and Gramfort, Alexandre and Michel, Vincent and Thirion, Bertrand and Grisel, Olivier and Blondel, Mathieu and Prettenhofer, Peter and Weiss, Ron and Dubourg, Vincent and others},
  journal={the Journal of machine Learning research},
  volume={12},
  pages={2825--2830},
  year={2011},
  publisher={JMLR. org}
}

@incollection{rasmussen2003gaussian,
  title={Gaussian processes in machine learning},
  author={Rasmussen, Carl Edward},
  booktitle={Summer school on machine learning},
  pages={63--71},
  year={2003},
  publisher={Springer}
}

@inproceedings{wielitzka2017state,
  title={State and maximum friction coefficient estimation in vehicle dynamics using UKF},
  author={Wielitzka, Mark and Dagen, Matthias and Ortmaier, Tobias},
  booktitle={2017 American Control Conference (ACC)},
  pages={4322--4327},
  year={2017},
  organization={IEEE}
}

@inproceedings{wielitzka2018sensitivity,
  title={Sensitivity-based road friction estimation in vehicle dynamics using the unscented kalman filter},
  author={Wielitzka, Mark and Dagen, Matthias and Ortmaier, Tobias},
  booktitle={2018 Annual American Control Conference (ACC)},
  pages={2593--2598},
  year={2018},
  organization={IEEE}
}

@inproceedings{wielitzka2019credibility,
  title={Credibility of state and friction coefficient estimation in vehicle dynamics using UKF},
  author={Wielitzka, Mark and Ortmaier, Tobias},
  booktitle={2019 IEEE 58th Conference on Decision and Control (CDC)},
  pages={3922--3927},
  year={2019},
  organization={IEEE}
}

@article{yoon2016estimation,
  title={Estimation of vehicle sideslip angle and tire-road friction coefficient based on magnetometer with GPS},
  author={Yoon, J-H and Eben Li, S and Ahn, C},
  journal={International journal of automotive technology},
  volume={17},
  number={3},
  pages={427--435},
  year={2016},
  publisher={Springer}
}

\end{document}